# Fixation Data Analysis for High Resolution Satellite Images

Ashu Sharma[1], Jayanta Kumar Ghosh[1], Saptrarshi Kolay[2]
1. Civil Engineering Department, Indian Institute of Technology, Roorkee, India
2. Department of Architecture & Planning, Indian Institute of Technology, Roorkee, India

**Abstract**
The presented study is an eye tracking experiment for high resolution satellite (HRS) images. The reported experiment explores the Area Of Interest (AOI) based analysis of eye fixation data for complex HRS images. The study reflects the requisite of reference data for bottom up saliency based segmentation and the struggle of eye tracking data analysis for complex satellite images. The intended fixation data analysis aims towards the reference data creation for bottom up saliency based segmentation of high resolution satellite images. The analytical outcome of this experimental study provides solution for AOI-based analysis for fixation data in complex environment of satellite images and recommendations for reference data construction which is already an ongoing effort.

**Introduction**

Advent arrival of High-Resolution Satellite (HRS) imagery data has rendered many civil engineering as well as other fields of applications such as surveillance, emergency response, 3D GIS and drone-based applications easier and economical. Therefore, a surge has been noticed in the recent years for the automatic interpretation and analysis of HRS images. Despite several years of research, complex and ambiguous image analysis is still an open problem, especially, when compared to human prowess for complex visual stimuli analysis. Because of multiple objects, HRS images of the urban area possesses complex composition. Thus, inculcating human prowess is required to analyse complex HRS image autonomously.

Human vision begins with visual search which is guided by attention or saliency. Salient objects form the foreground which separates them from background. Likewise, for HRS image segmentation, which is a key step in image analysis, saliency is a popular subject of interest. [Zhang et al 2015, Diao et al 2016, Li & Itti 2011, Sirmacek et al 2013, Zhu et al 2015, Zhai et al 2016, Li et al 2016]. But, as final image analysis depends over the underlying segmentation method, a reasonable and adequate assessment for the segmentation method is also necessary. In case of non-complex HRS image or top down saliency based HRS image segmentation, the reference data creation is straightforward and simple. For non-complex HRS image, the delineated boundary of conspicuous object/s serves as reference [Zhang et al 2015, Diao et al 2016], whereas in top down saliency based segmentation, the target object/s form foreground which are used as ground truth [Li & Itti 2011, Sirmacek et al 2013, Zhu et al 2015, Zhai et al 2016, Li et al 2016].

But, due to the presence of multiple objects in complex/cluttered HRS images, the conspicuous object/s identification is vague. Moreover, predefining a unique salient object/s for bottom up saliency based segmentation (BSS) of complex images is perplexing. This renders the creation of a standard ground truth (GT) for assessing BSS of complex HRS images. Also, as bottom-up saliency output serves as input to top down systems and always dominates over top-down [Frintrop 2011, Ramanathan et al 2010], studying BSS of complex HRS images is an issue of huge importance.

In this regard, human search pattern and inner psychovisual process can guide for the standard reference data solution for complex HRS image segmentation. The seminal work by Guy T. Buswell 1935, Yarbus 1967 demonstrated the relationship between eye movements and cognitive processes. Therefore, along with training a model to predict the human like fixations

[Tilke Judd 2009], the eye tracking data provide an excellent modality to assess saliency based segmentation for complex images [Jacob et al 2003, Koostra et al 2011]. Also, the region based saliency models, which are well suited for image segmentation, highly relies over eye tracking dataset. [Achanta et al 2008]. Fixation points from eye tracking data provide the precise salient seed locations for segment generation [Mishra et al 2009, Ramanathan et al 2010, Tavakoli Thesis 2014]. The area (or object/s) with high fixation density becomes foreground and rest serves as background and thus, used as reference. Although, some datasets with eye tracking based GT are already available for indoor/outdoor and terrestrial images [Judd et al 2012, Hou et al 2007]. But, absence of eye tracking based GT dataset for BSS of complex HRS images motivates the reported experiment.

Although, the readymade data products from eye tracker, such as heat maps and gaze plots, provide different types of data visualization. But as human see this world in terms of crisp functional categories and objects [Freedman et al 2002], heat maps do not provide a direct assessment for the same. Other data products, such gaze plots, are not suitable for large set of data due to multiple overlapping. Although they are good for comparing two scan paths [Andrienko et al 2012], but are not ideal for generalized solution. Therefore, for final FBS like GT, processing of fixation data is needed to establish a relation between human visual attention and salient object/s of complex visual stimuli [Jacob et al 2003].

In this way, Area Of Interest (AOI) based eye tracking data analysis facilitates for delineating object/s and assigning weightage according to the fixations [Kurzhals et al 2017]. AOI-based analysis of eye tracking data and AOI-based metrics provide an easy way for gaining the insight about the human tendency towards visual stimuli. In the case of top-down study, the AOI can be easily drawn by using the target object and according to the given task, the "events" can be created [Kurzhals et al 2017]. But in case of BSS of complex stimuli, presence of multiple objects in occluded manner arises perplexing situation for drawing AOI. Also, as fixation doesn't mean recognition, the combination of objects recognised through 'think aloud method' and heat maps doesn't solve the core problem of reference data creation. Also, the current literature lacks any standard AOI based task analysis method for bottom up studies in complex environment. Therefore, the objective of this study is to provide a solution for AOI based eye tracking data processing in order to prepare the GT data for BSS of complex HRS images, which is currently an ongoing effort.

**Literature Survey**
According to the objective, this literature survey sums up the study about dataset for satellite images, eye tracking based GT and different methods of AOI based task for eye tracking data analysis. The study of different datasets provides the insight into the existing solutions for different types of standard and well-accepted references in the field of image segmentation or object detection. As any analytical study is incomplete without its validation and performance assessment, a standard benchmark is always a sine-qua-non. In case of saliency based segmentation for non-complex images numerous benchmarks along with eye tracking data are freely available online [Bruce, Judd, Cerf, FT, IS]. Among them, MSRA presents salient object/s annotation in terms of rectangle boundary box for a large image dataset (5,000 images) with smooth background [Liu et al 2007]. On the other hand, MSRA-1000 [Achanta et al 2009] and MSRA10K [Cheng et al 2013], subsets of the MSRA dataset, provide pixel-wise accurate ground truth. But neither of these datasets provide salient object analysis in complex images nor do they have satellite images in their sample domain.

For complex images, though Extended Complex Scene Saliency Dataset (ECSSD) [Shi et al 2016] uses complex and cluttered background images, but does not contain multiple salient objects. In the same scenario, DUT-OMRON dataset, provide boundary box for one or more salient objects along with eye fixation data in relatively complex background [Yang et al 2013]. But this attempt does not fulfil the need for pixel-wise object annotation and satellite imagery domain. NUSEF, another eye tracking-based reference data exhibits the dominance of image content. The dataset consists of visual stimuli from many semantic categories, including images with complex composition. [Ramanathan et al 2010]. Li et al 2013 offers the pixel based ground truth and fixation data (PASCAL-S), where images are collected form PASCAL 2010 dataset to explore the connection between fixation points and salient object segmentation. The PASCAL-S dataset assigns different saliency value for multiple salient objects by using eye tracking fixation data. This is the only dataset which provide benchmark for multiple salient objects annotation in an image, altogether. This study also discusses dataset design bias. Analogically, presence of multiple objects in HRS images renders the similar approach suitable for BSS benchmarking. Although, these many benchmarks are available for salient object segmentation but none of them uses satellite image in their sample image domain.

CAT 2000 is the only dataset where one category contains eye tracker based saliency maps for satellite images [Borji el al 2015]. Though, these maps are good for training any algorithm for fixation point generation but do not provide solution for final (object/s) segmentation or FBS assessment. Also, eye-tracking data are not general in nature and varies for multiple reasons [Jiang et al 2015]. Therefore, direct combination of different datasets is not possible nor any model learned from one dataset directly generalize to another. Another benchmarks, such as RSSCN7 Dataset [Zou et al 2015], UC Merged Dataset [Yang and Newsam 2010] and WHU-RS19 [Dai and Yang 2011, Sheng el al 2012], provide category based ground truth for satellite images, where the images are specific and conspicuous object/s are based on the predefined category. Thus, lack of benchmark dataset for BSS of HRS images and use of eye tracker in many of the other benchmark datasets propel eye tracking experiment for fixation data analysis in bottom up fashion for HRS images.

Keeping all the above in view, an eye tracking based bottom up study for HRS images has to be carried out. There are different analysis methods to analyze the eye tracking data [Andrienko et al 2012, Blascheck et al 2014, Blascheck et al 2016, Kurzhals et al 2017]. These methods are used to derive three independent data dimensions and can be broadly classified as: 1. Time based analysis, 2. AOI based analysis and 3. Participant's category based analysis. The Time based analysis methods give the 'When' of the first fixated areas of image. Whereas AOI based analysis typically answers the 'What' and 'Where' of the fixated area. Participant's based analysis is dependent on the participant's background and category. This answers the certain viewing behaviour and helps in mining the underlying thinking strategy of different people. Generally, various measurements for eye fixation data are obtained on the basis of AOI information. AOIs are drawn on the basis of object boundary and different metrics such as 'fixation count', 'first fixation', 'visit counts', 'number of fixations' etc. are calculated [Andrienko et al 2012, Kurzhals et al 2017]. Accordingly, the AOI with maximum fixation serves as the foreground. In some other eye tracking experiments, corresponding to the fixations, a ranking is assigned for multiple objects AOIs for FBS [Li et al 2014]. Some existing eye tracking studies for HRS images also include the AOI based analysis which are based on the predefined intended task [Matzen et al 2016, Dong et al 2014]. Although, heat maps are direct representation of eye fixations and good for data visualization but they are not appropriate for quantitative analysis. Moreover, for benchmark creation a generalized and quantitative standard salient segmented area is required, hence, AOI based analysis seems the most suitable for fixation analysis of HRS images.

**Problem Definition and Eye Tracking Experiment**
In spite of the existing eye tracking studies, AOI creation in bottom up fashion is perplexing. Unlike Li et al 2013, sometimes, due to occlusion or other reasons, objects are not disjoint in HRS images. Also, as fixations are often located on or outside the boundary of object AOI [Ramanathan et al 2010, Lee & Lee 2015], a careful drawing of AOI is of significant importance. In addition to that, fixation points are in overlapping fashion and cover multiple objects at a same time, while all of those objects may not be consciously perceived. This issue is described and discussed in a later section. Consequently, the problem is defined as drawing AOIs for eye tracking data analysis with respect to the conscious recognition of that AOI as object or object category. Considering all above, the eye tracking experiment is carried out and particulars of the experiment are given below.

*Participants*
Total 38 participants (27 male and 11 female) volunteered from IIT Roorkee. The participants have normal or corrected to normal vision and lie in age group of 22-35 years. The participants are recruited from two categories; one is expert level who have geomatics background and are good with satellite image interpretation. Second category is of user level who are familiar with satellite images but cannot interpret an image as an expert. The former category have 14 participants and rest belong to the latter category. In order to make the experiment completely unbiased, no participant knew in advance about the dataset to be shown. Although, the study aims the bottom up saliency based analysis, but the participants are, at least, familiar with satellite images. Due to this situation, the experiment is slightly biased with this pre-learning.

*Stimuli*
There are total 27 RGB HRS images. These images are different non-overlapping cropped areas of 5 sample images. Among them two sample images are of Worldview-2 satellite and are of Washington, D.C. (June 8, 2011) and Madrid, Spain (February 7, 2011) area. Another two sample images are of Geoeye-1, (area Taj Mahal, Agra, India; dated October17, 2009) and Vatican City, (Italy Rome; dated April 24, 2011). Both Worldview-2 and Geoeye-1 sample images have a spatial resolution of 0.50-meter. One more sample image is of Worldview-3, (area Paris Pont Des Arts Bridge) with the 0.30-meter spatial resolution, finest among the rest of sample images. The images are of different size and pixel resolution varies in between 220 X 260 to 625 X 560. No pre-processing has been done to the images before showing to the participants.

*Experimental Setup*
Stimuli are shown on the 52 X 32.57 cm screen having 1920 X 1200 pixel resolution. The distance from the screen to participant always lies in between 65-75 cm. The images are shown for 6 seconds with a blank screen in between each pair of images for 2 seconds to mitigate priming effect. For clear object recognition, images are shown with scaling. As the eye tracker used is glasses, this scaling does not affect the real fixation mapping on respective images.

*Eye Tracker Instrument and data Acquisition*
The Tobii Pro Glasses 2 wearable eye tracker is used for data collection. The instrument has been calibrated for each participant by using calibration card provided along with instrument by keeping in between the distance of 0.75 and 1.25 meters. The wearable technology gives the participants the freedom to view the stimuli in more natural way. All participants are asked for free viewing. Other than "think aloud", no prior instruction is given before "free viewing" task.

**Observation and Discussion**
*Observation*
All objects recognized through "think aloud" process are listed and accordingly objects in an image are defined for the frequency of participants. Sometimes, instead of discrete object recognition, some participants have tried to describe the image. Hence, frequency of participants for 'Description' is also listed, though sometimes the guess is not correct. Table 1 shows the frequency of participants for conscious recognition of respective objects. For the sake of ease, Table 1 consists data for only 3 images from dataset which are further used to describe the results.

Table 1: Objects Recognized During 'Think Aloud' Process

| Image 1 | | | | | | |
|---|---|---|---|---|---|---|
| Objects | Road | House/ Building | Trees | Cars | Ground/ Plaza/ stadium | Description |
| Number (#) of participants | 7 | 24 | 17 | 3 | 4 | 18 |
| Image 2 | | | | | | |
| Objects | Bridge | Water/ Canal/ River | Trees | Building | Car/ Vehicles | Road | Description |
| #of participants | 26 | 19 | 5 | 13 | 6 | 5 | 6 |
| Image 3 | | | | | | |
| Objects | Park/ Garden/ Lawn | Monument | Built-up Area | Small Roads | Trees | Description |
| #of participants | 23 | 24 | 11 | 1 | 1 | 8 |

*Objects-Based AOI Drawing:*

Recorded eye tracker data has mapped without any filter for each image. Now for metric analysis AOI has to be drawn. Initially irrespective to the object category, discrete AOIs are drawn for all presented objects; except the small ones such as cars or very small trees. The AOI for three sample images along with original images are given in Fig. 1. Three metrics namely 'First Fixation' (FF), 'Fixation Duration' (FD), and 'Fixation Counts' (FC) are measured and results are matched with the observation given in Table 1. Their corresponding metric plots are given in Fig. 2, Fig. 3, and Fig. 4 for FF, FD and FC respectively where 'n' in legend is the number of participants. All of the metrics are plotted by using raw fixation data.

While analysing the data, an incoherence is observed between AOI-based metrics and the consciously perceived objects. E.g. in image 2, according to the objects, road and bridge are two different objects. Here metrics boast about road with early fixation, high duration and more counts. Whereas most of the participants identified 'bridge' as an object, not the road (identified by very less number of participants). The similar conflict has been observed all over the dataset. Some examples given in Table 1 and Fig. 1-4 reflects the same incoherence conspicuously. As mentioned earlier under problem definition, one possible reason for this can be as mentioned by [Ramanathan et al 2010, Lee & Lee 2015]. In that case, most of the fixations will not lie inside of the objects' boundary. The same can also be perceived from the data visualized through heat maps (Fig. 1) and object recognized (Table 1). Furthermore, the objects in the centre gets more fixation as compared to others but need not to be recognized or attended by most of the participants. E.g. in image 1, 'trees' are in centre and therefore metrics have high count for it; whereas less than half participants recognized 'trees' that too at later point of time. The possible reasons may be low color contrast, or some psychological thought process which varies from person to person. Thus object-based AOIs drawing does not provide any insight about the psychovisual factors for conscious object recognition and therefore, cannot be used for further GT creation.

However, object attended (eye movements) and object recognized are apparently co-related. Therefore, there has to be some certain way to draw AOI for digging out the inner psychovisual factors. To understand the same, certain points are needed to be considered. First, Eye tracker captures only the foveal region where the vision is clear, whereas humans do perceive in the parafoveal and peripheral region as well [Ramkumar et al 2012, Pannasch et al 2011]. In both of the regions edges are formed with contrast and impression of object has been sensed. Moreover, the overall impression of the visual scene is typically perceived within first few seconds of exposure [Ramkumar et al 2012, Pannasch et al 2011]. E.g. there are multiples of buildings in image 1 and participants also recognize "building" as a dominant object. But for recognizing "many buildings" participants need not to fixate on each of the building. Rather, separately, these buildings are fixated at a later point of time with less fixation duration and count as compared to other objects (Fig. 2(a), 3(a) & 4(a)).

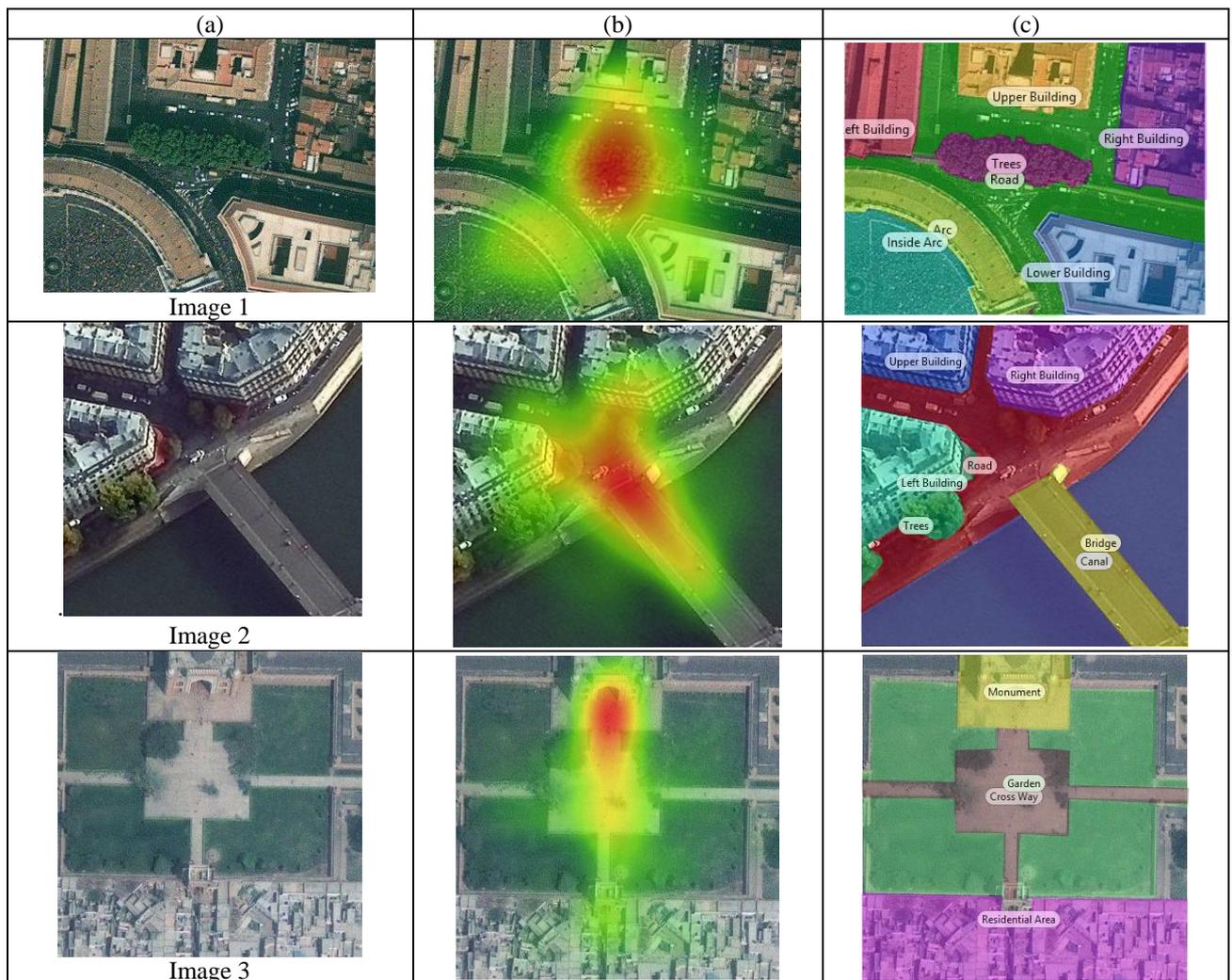

Figure 1. (a) Original images, (b) heat maps, (c) Object based AOIs drawn

Moreover, in case of satellite image, since all participants possess user level proficiency and no predefined task is given, the first impression as an aerial view of some geographical location is acquired by the participants. Because bottom up saliency is driven by the perceptual grouping of low level features of the object and always dominates, maximum participants consciously recognize/identify objects. On the other hand, due to the previous learning (user level proficiency) of participants, perceived height, color, shape and other factors affects the attention [Lang et al 2012] and perceptual grouping occurs in form of functional grouping. The same is observed during "description" given by participants. Thus, it is inferred that due to the previous basic learning, eye movement behaviour during attentional search is guided by functional grouping and conformably, participants attempt to identify objects. To examine this hypothesis, AOI based task analysis is moulded with functional grouping and metrics have been compared with the conscious object identification. The separate discussion for the same is given in the following section.

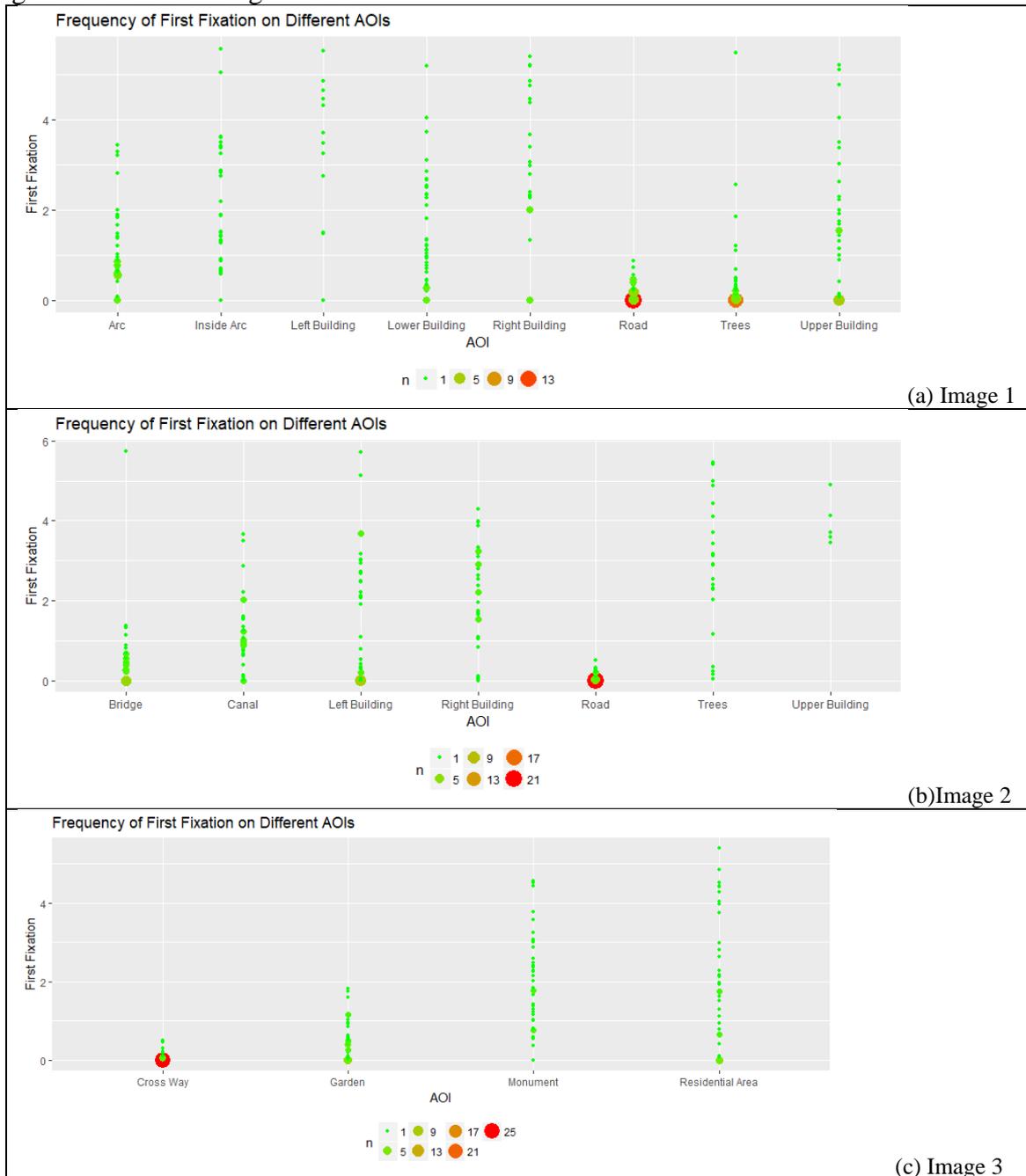

(a) Image 1
(b) Image 2
(c) Image 3

Figure 2. First Fixations

(a) Image 1

(b) Image 2

(c) Image 3

Figure 3. Fixation Duration

(a) Image 1

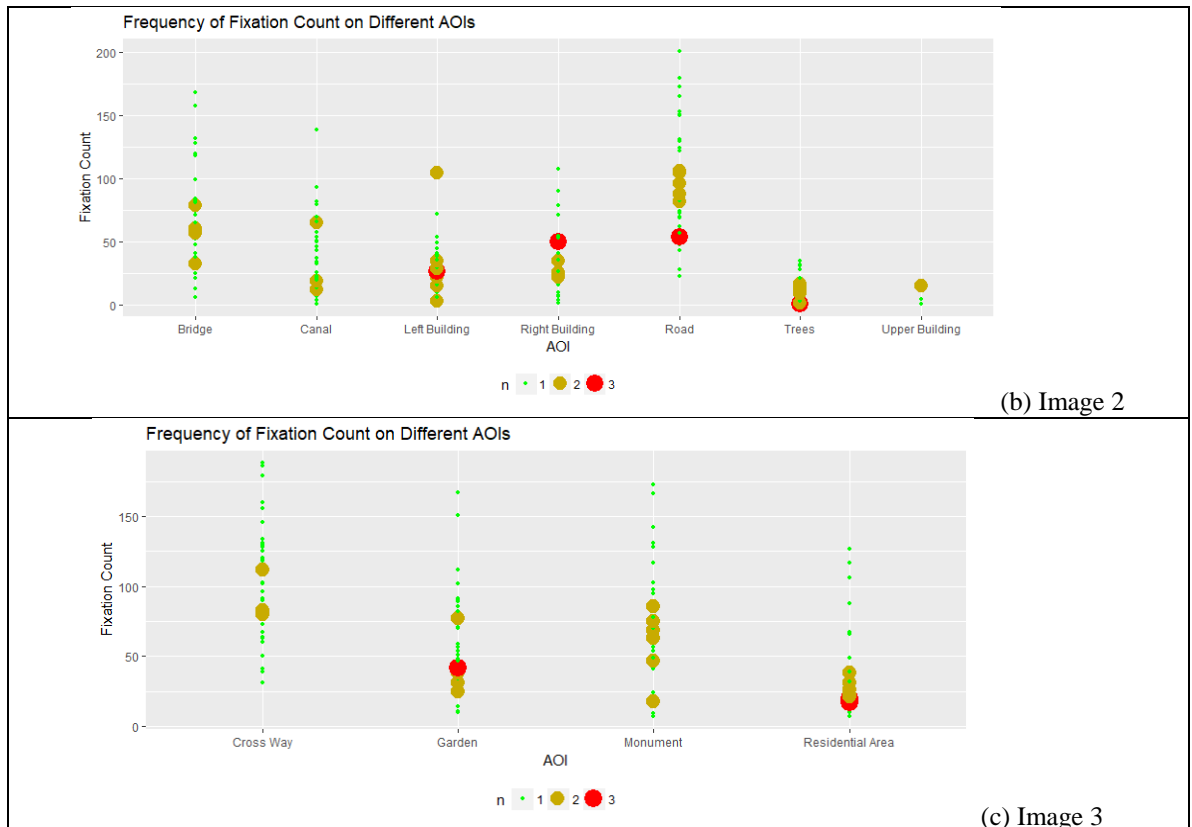

Figure 4. Fixation Count

*Functional Grouping-Based AOI Drawing:*

Now, for functionally grouped AOIs creation, possible layers have been generated and analysed. The AOIs drawn for example images have been shown in Fig. 5 and Fig. 6-8 represents the corresponding metrics. Objects falling within same functional category have collectively included in a single AOI and likewise layers have been generated. While drawing AOIs, small discrete object such as cars, small trees have been discarded form AOI generation. Some examples for AOI drawn are: in image 1 'Road' and area 'Inside Arc' possess different functional category, therefore, separate AOIs have been drawn. Whereas, in image 2 'Road' and 'Bridge' have same functional; hence single AOI is drawn for both. Also, while perceiving the functional grouping, 'Gestalt principles for perceptual grouping' also plays a vital role. For example: in image 3 'Garden' and 'Crossway' are on same height with different functional group, but 'common region and element connectedness' renders the four parts of 'Garden' as one which are divide by 'Crossway' [Richtsfeld et al 2014, Rock and Palmer 1990, Palmer 1992, Palmer and Rock 1994]. Therefore, single AOI for entire region and the 'crossway' is considered as the part of garden not as separate road. Here one thing should be noted that as the intended aim is to create a standard reference data for BSS, these AOI should be drawn in such a way that can further lead towards identification and recognition of objects.

While testing the inferred hypothesis for functionally grouped AOIs; metrics and objects recognized during think aloud are found to be co-related. E.g. in image 2, 'Road and Bridge' AOI is having high fixation duration with high fixation count and the same is also reflected from what participants have consciously recognized. Whereas, previously metrics boasted for 'Road' and not for 'Bridge'. Similarly in image 1, AOI drawn for 'Buildings' as a single functional group have comparatively high fixation duration and count. Here, saliency has shown a relevance and relation while assigning the name for the attended AOI. Continuing with previous example of image 2, the most attended AOI is 'Road and Bridge' and most spoken object during think aloud is 'bridge'. Here, 'bridge' object is more salient than 'road' and therefore the most attended AOI is considered as 'Bridge'. Analogically, in image 1, though metrics supports for 'Road' AOI but 'road' behaves as background and therefore, is less salient. Although, the central biasing is a common biological propensity of human visual system [Koostra et al 2011], due to which, the metrics do exhibits the high biasing towards the central region of images. For the same reason in image 1 'Trees' have high fixation duration and count but less than half participants said 'tress' during think aloud.

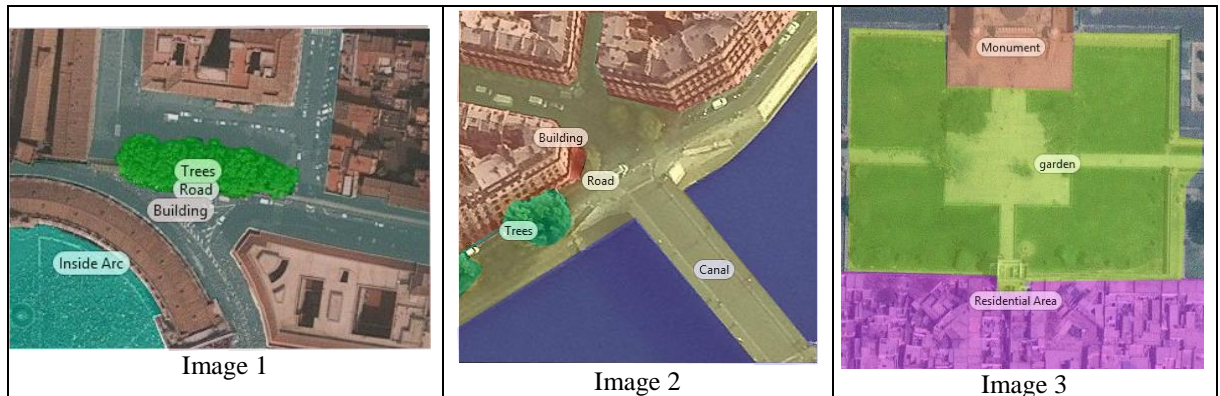

Figure 5. New AOIs drawn for Images

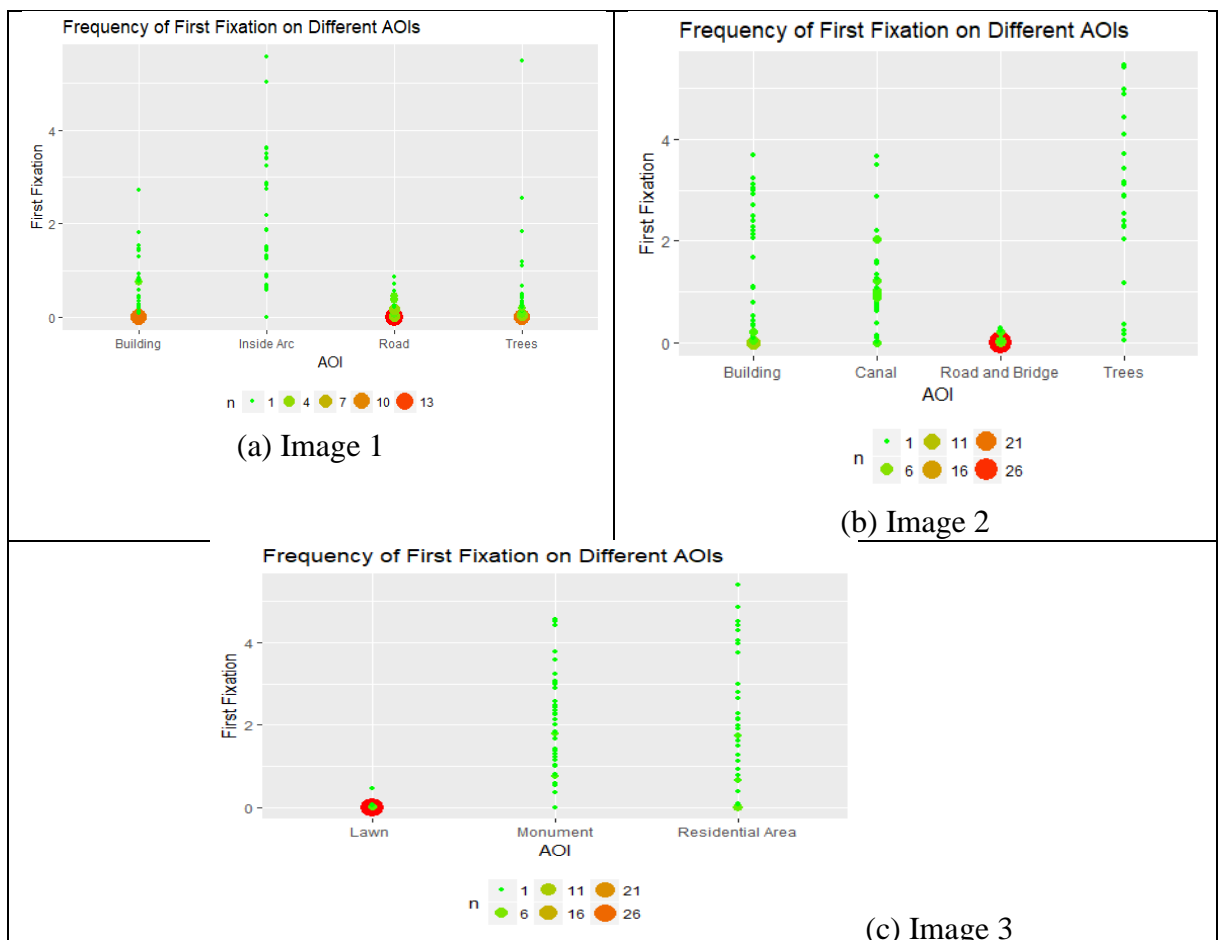

(a) Image 1

(b) Image 2

(c) Image 3

Figure 6. First Fixations

Figure 7. Fixation Duration

Figure 8. Fixation Count

*Overall Discussion:*
From the above observation of the experiment, relevance of perceptual grouping is quite evident which influences the eye movements and attention. Previous learning of participants enables the top view perception of geographical location. Edges sensed in entire visual region (in foveal, parafoveal and peripheral regions) forms the shape of objects and along with other interpretation elements, such as perceived height, color, etc, the object is identified. Affordance [Gibson 1979] offered by each objects is intervened by perceptual grouping and finally layers are perceived which are made by functional group of objects. The metrics for functionally grouping- based AOI supports this conclusion.

While the central tendency of vision is ignored, influence of saliency has been noticed in adjacent functionally grouped layers. These adjacent functionally grouped layers compete for saliency with each other and more salient object/layer is identified consciously. During the recognition of objects, multiple foreground and background (Figure and Ground) are formed and according to the degree of saliency, order of object recognition take place. E.g. in image 2, primarily, there are three layers for which AOIs are drawn- 'Buildings', 'Road and Bridge' and 'Canal'. For the sake of ease, say name of these layers are A1, A2 and A3 respectively. Now, in between A1 and A2 layers; 'road' serves as the background while 'buildings' becomes the foreground. On the other hand, in between A2 and A3; 'bridge' acts as foreground (more salient) while 'canal' serves as background. But, in between the adjacent layers of 'A1 and A2'; and 'A2 and A1'; 'bridge' with respect to 'Canal' layer is more salient than the 'road' with respect to 'Buildings'. Here one this is to be noted that although road and bridge are a single functionally grouped layer altogether, but while comparing, 'road' object is compared with 'Building' layer; not the entire A2 layer AOI. Similarly, this happens with 'Canal' layer while comparing with 'bridge' object. Also, while striving, visual features like contrast, aesthetics, geometric structures etc. gain more attention [Liu et al 2012]. Thus, during multiple figure-ground segregation, 'bridge' is the most consciously recognized and dominant object in image 2 and therefore, said by maximum participants during think aloud process. The rationale behind this phenomena can be justified by attentional selection and crowding phenomena [Lamme 2003]. Because of which, visual features renders the evolution of perceptual organization and among multiple interaction of these groups only few reaches the conscious visual experience at later point of time.

If central tendency of vision is considered, centre or near centre fixation occurrence is observed in a way that retains 'robustness of shape' [Ommer 2013, Ghosh and Petkov 2005]. Not only, but different factors for saliency, as mentioned above [Liu et al 2012], also influences these fixations. For example in image 2, maximum central region is fixated where there is no building and yet, more than half of the participants consciously recognized 'building' as an object (Table 1). Once the objects are perceived as a single functional group, form perception in the whole visual region enables the freedom to fixate partially and not the each building within the 'Building' layer. Here, if a part of a functional group is recognized then the each object in that group/layer will also be identified. Now, saliency due to the geometry of the arc shape building and contrast of lower right building in 'Building' layer drags more attention (Fig. 1(b)). Therefore, near centre part of these building attain more fixations and thus, with average fixations (Fig. 6-8 (a)) all buildings are identified. Although, some fixations also happen at the top middle building, possibly, which is fixated for retaining the robustness of shape of entire 'Building' layer. Although, the heat map analysis is out of scope of this reported study, yet the rationales during such discussions provide the concrete for the theoretical foundation to form guidelines for intended reference data generation.

Also, from this AOI-analysis, dominance of perceptual organization over the functional categorization is observed. E.g. in image 3, four parts of 'Lawn' are perceived altogether; whereas lawn and crossway have separate functional categories and should be recognized separately. Furthermore, in spite of central fixations, 'crossway' is considered as a part of garden and therefore, very few participants said the 'crossway' separately. The occurrence of perceptual grouping in absence of conscious vision [Montoro et al 2014] can be the possible rationale behind such preference. Although, the functional grouping requires some previous learning and cannot be done completely in bottom up manner.

**Conclusion and Future Scope**

From the experiment observation and discussion for AOI-based fixation data analysis, some important conclusions can be drawn, which can be useful while building the reference data for BSS. First, while analysing the fixation data for complex HRS images in bottom up manner, functional grouping based AOIs are more relevant and related to what humans will actually perceive. Second, while drawing the AOIs, perceptual grouping should be given preference over the functional grouping. And finally third, the metrics value calculated for functionally grouped AOIs varies with the degree of saliency for each AOI layer which is identified by comparing the adjacent layers. More salient object or part of AOI layer is considered to be the most consciously recognized object and hence, possesses the highest values for the metrics. Although, irrespective of any factor, central or near central region always have high fixations.

Thus, from the above conclusions, it can be inferred that for bottom up saliency based studies, the reference data should contain multiple layers where the objects are functionally grouped. And according to the assessed saliency, some degree should be assigned to each layer, where central layers will always have high weighting.

Hence, the presented study suggests the 'How' of AOI-based analysis of eye tracking data in complex environment for satellite images. The presented AOI based task analysis is able to dig insight and measure the quantitative metrics which provides aid in understanding the inner psychovisual factors. Though, the central tendency always affect the final metric analysis but, it can be handled by analysis of comparative saliency for adjacent layers.

Future scope of this experiment has vast region and useful in many eye tracking studies. Reference data creation which is in progress in already an example of it.